\documentclass[fleqn,10pt]{wlscirep}
\usepackage[utf8]{inputenc}
\usepackage[T1]{fontenc}
\usepackage{graphicx}
\usepackage[dvipsnames]{xcolor}
\usepackage{dirtree}
\usepackage{subcaption}
\usepackage{caption}
\usepackage{cleveref}
\usepackage[export]{adjustbox}
\usepackage{hyperref}
\usepackage{threeparttable, tablefootnote, siunitx, booktabs}

\newcommand{\orcid}[1]{
    \href{https://orcid.org/#1}{
        \includegraphics[height=2ex]{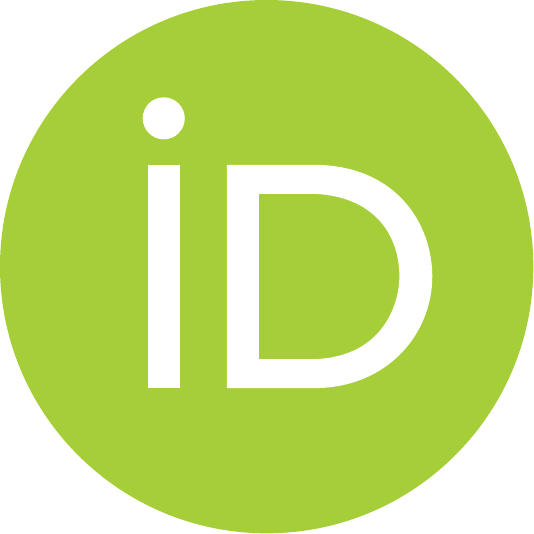}
    }
}
\usepackage{lineno}

\title{Fluorescent Neuronal Cells v2: Multi-Task, Multi-Format Annotations for Deep Learning in Microscopy}

\author[1,2*]{Luca Clissa}
\author[3]{Antonio Macaluso}
\author[2]{Roberto Morelli}
\author[4]{Alessandra Occhinegro}
\author[4]{Emiliana Piscitiello}
\author[4]{Ludovico Taddei}
\author[4]{Marco Luppi}
\author[4]{Roberto Amici}
\author[4]{Matteo Cerri}
\author[4]{Timna Hitrec}
\author[1, 2]{Lorenzo Rinaldi}
\author[1, 2]{Antonio Zoccoli}
\affil[1]{National Institute of Nuclear Physics, Bologna, Italy}
\affil[2]{University of Bologna, Department of Physics and Astronomy, Bologna, Italy}
\affil[3]{German Research Center for Artificial Intelligence (DFKI), Agents and Simulated Reality Department, Saarbruecken, Germany}
\affil[4]{University of Bologna, Department of Biomedical and Neuromotor Sciences, Bologna, Italy}

\affil[*]{corresponding author(s): Luca Clissa (clissa@bo.infn.it)}

\begin{abstract}
Fluorescent Neuronal Cells v2 is a collection of fluorescence microscopy images and the corresponding ground-truth annotations, designed to foster innovative research in the domains of Life Sciences and Deep Learning.
This dataset encompasses three image collections in which rodent neuronal cells' nuclei and cytoplasm are stained with diverse markers to highlight their anatomical or functional characteristics.
Alongside the images, we provide ground-truth annotations for several learning tasks, including semantic segmentation, object detection, and counting.
The contribution is two-fold.
First, given the variety of annotations and their accessible formats, we envision our work facilitating methodological advancements in computer vision approaches for segmentation, detection, feature learning, unsupervised and self-supervised learning, transfer learning, and related areas.
Second, by enabling extensive exploration and benchmarking, we hope Fluorescent Neuronal Cells v2 will catalyze breakthroughs in fluorescence microscopy analysis and promote cutting-edge discoveries in life sciences.
The data are available at:
 \href{https://amsacta.unibo.it/id/eprint/7347}{https://amsacta.unibo.it/id/eprint/7347}.
\end{abstract}
\begin{document}


\flushbottom
\maketitle

\thispagestyle{empty}


\section*{Background \& Summary}


Fluorescence microscopy is as a pivotal imaging technique in life-science experiments, allowing researchers to study biological structures or processes with remarkable precision. It employs fluorescent dyes or proteins that emit light at specific wavelengths depending on the illuminating wavelength they absorb. 
Exploiting this phenomenon, specific molecules can be tagged (\textit{stained}) with fluorescent markers, and visualized by filtering only their emitted light, thus providing valuable insights into their localization, activity, and interactions.

Despite its widespread use, current practices in fluorescence microscopy analysis heavily rely on semi-automatic procedures, often necessitating manual recognition and/or counting of specific neuronal structures of interest \cite{luppi1, luppi3,chiocchetti2021phosphorylated}.
For instance, in the study of torpor mechanisms, researchers depend on laborious hand-crafted operations to identify neuronal networks associated with this process \cite{hitrec2019neural}. This manual aspect typically delays the analyses, also introducing potential errors due to limitations of human operators. Moreover, the similarity between structures of interest and the background often leads to challenges in distinguishing and accurately recognizing biological compounds, resulting in inherent arbitrariness and interpretation bias.

For these reasons, there is a growing interest in automating the recognition and counting of tagged elements in fluorescence microscopy\cite{morelli2021cresunet,cao2020denseunet,Riccio2019,kumar2020multisegmentation}. 
Deep learning approaches have demonstrated great promise in various object recognition tasks. However, their performance can deteriorate when applied to data from domains significantly different from those adopted for pre-training (\textit{domain shift}\cite{medical_domain_shift,poon2023dataset}). Furthermore, the effectiveness of these approaches typically relies heavily on the availability of well-annotated data\cite{curse_dataset_annotation}, which is often scarce and limited in the fluorescence microscopy domain.

To mitigate these issues,
we present the Fluorescent Neuronal Cells v2 (FNC) dataset. This archive features 3 data collections, for a total of 1874 high-resolution images of rodents brain slices capturing a diverse range of neuronal structures and staining patterns. To facilitate research in this field, we also provide 750 annotations in various formats, tailored to popular supervised learning tasks such as semantic segmentation, object detection, and counting. 
Apart from serving as an additional benchmark for testing model generalization in microscopy applications, the FNC dataset opens up several research opportunities.
Firstly, the heterogeneity of biological structures and their visual characteristics enable testing the generalization of trained models, and validating transfer learning and domain adaptation methods \cite{haq2020adversarial_domain_adaptation,brieu2019domain_adaptation}. 
Also, the availability of multiple annotation types allows the exploration of different learning paradigms, ranging from supervised and unsupervised approaches to self-/weakly-supervised techniques. 
Moreover, the specific challenges of our data well suit investigations into methodological advancements, e.g., assessing the effectiveness of different annotation formats and uncertainty estimation.

The design of the data collection process involved two distinct stages. Firstly, data collection was conducted following standardized experimental protocols. Specifically, controlled experimental conditions were applied to the animals, whose brains were sliced and processed by a classical immunofluorescence protocol to stain various neuronal substructures.
Subsequently, a fluorescence microscopy was employed to capture high-resolution images of the areas of interest.
Secondly, domain experts performed data annotation providing ground-truth labels necessary for supervised learning.

\begin{figure}
    \centering
    \includegraphics[width=\textwidth]{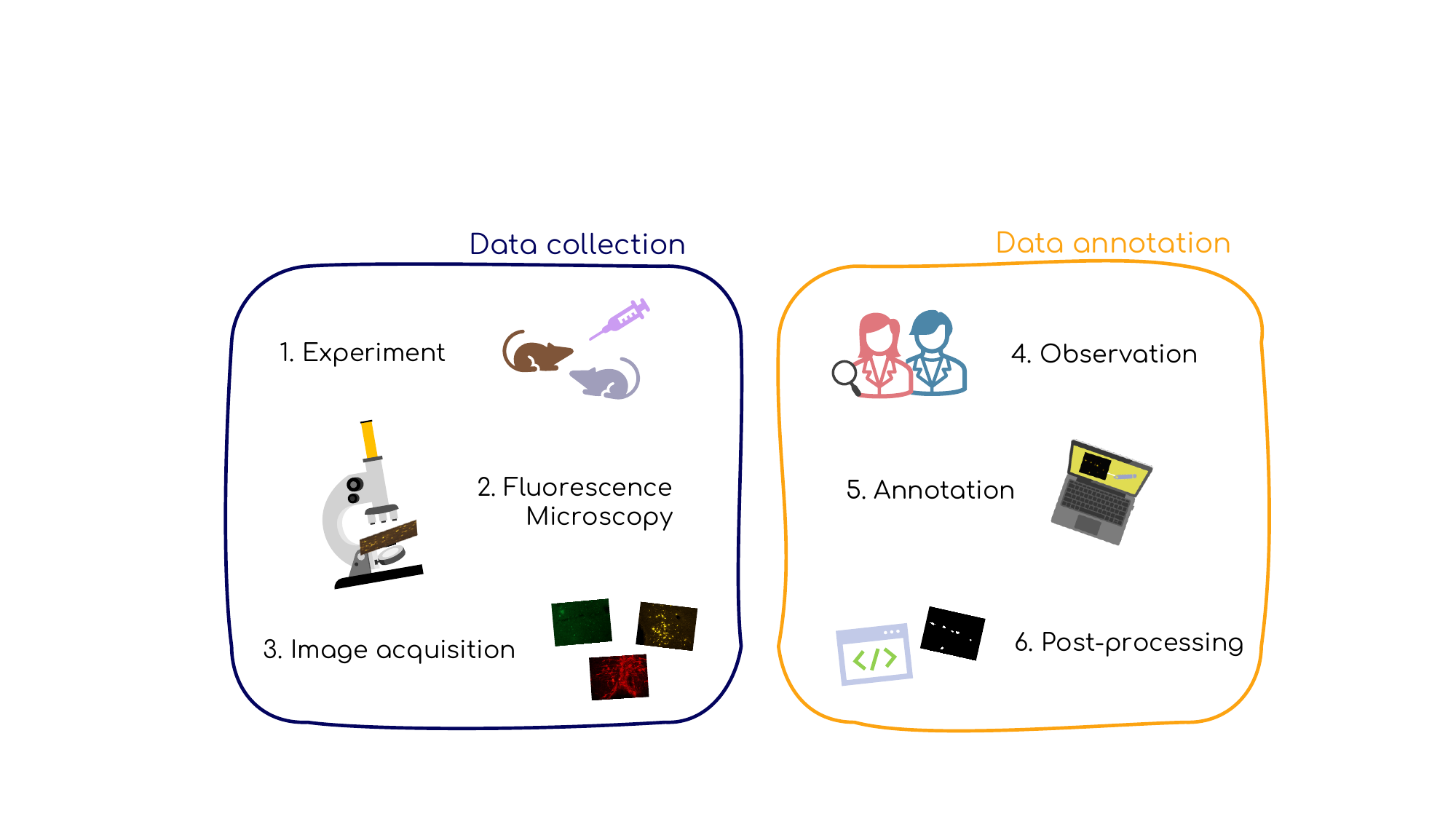}
    \caption{\textbf{Study design.} The study was designed in two phases: \textit{data collection}, where high-resolution pictures of rodent brain slices were acquired; and \textit{data annotation}, where expert researchers collected annotations needed for supervised learning approaches.}
    \label{fig:design-diagram}
\end{figure}

Despite the presence of open-source fluorescence microscopy datasets, several issues hinder their utilization for training deep learning models. 
Firstly, these collections typically lack accompanying ground-truth annotations, thus precluding the adoption of supervised learning techniques.
Secondly, labelled datasets often include just a few dozens of images\cite{RAZA2019160,SabineTaschner-Mandl2020}, that can be restrictive considering the data-intensive nature of deep learning models.
Also, the moderate resolution  of images in open datasets\cite{waithe_dominic_2019_2548493,stringer2021cellpose} hampers the effectiveness of resorting to crops as an alternative to whole images for augmenting sample size.
Thirdly, most existing datasets predominantly include a single marker type\cite{RAZA2019160,SabineTaschner-Mandl2020}, thus lacking diversity and limiting robust model training. 
Alongside these aspects, public datasets typically provide label types as dot-annotations or bounding boxes\cite{RAZA2019160,SabineTaschner-Mandl2020,waithe_dominic_2019_2548493,stringer2021cellpose}, which prevents their extension to fine-grained segmentation tasks.
Additionally, the data accessibility is sometimes restricted due to the use of domain-specific formats\cite{poon2023dataset}, which complicates integration into deep learning frameworks and wide dissemination.

In response to these challenges, we present a large archive comprising high-resolution fluorescent microscopy images, encompassing different markers and cell types. 
Furthermore, the data are shared as easily accessible \textit{PNG} files, and the corresponding annotations are provided in various types and formats, enabling the exploration of different learning approaches and tasks, thereby significantly expanding the scope of potential applications.


\section*{Methods}






 

The FNC dataset compiles images acquired from multiple studies and experimental conditions, while maintaining a consistent structure in the acquisition pipeline. Minor modifications were made to accommodate the specific requirements of each study and adapt to the current experimental circumstances and equipment.
The data collection process consisted of two distinct and independent stages: \textit{image acquisition} and \textit{data annotation}. This section provides a comprehensive description of the data acquisition design, including the dedicated measures implemented for each image collection (refer to Figure \ref{fig:design-diagram} for a visual summary).


    

\begin{figure}[htp]
    \centering

    \begin{subfigure}{.33\textwidth}
        \centering
        \includegraphics[width=\linewidth]{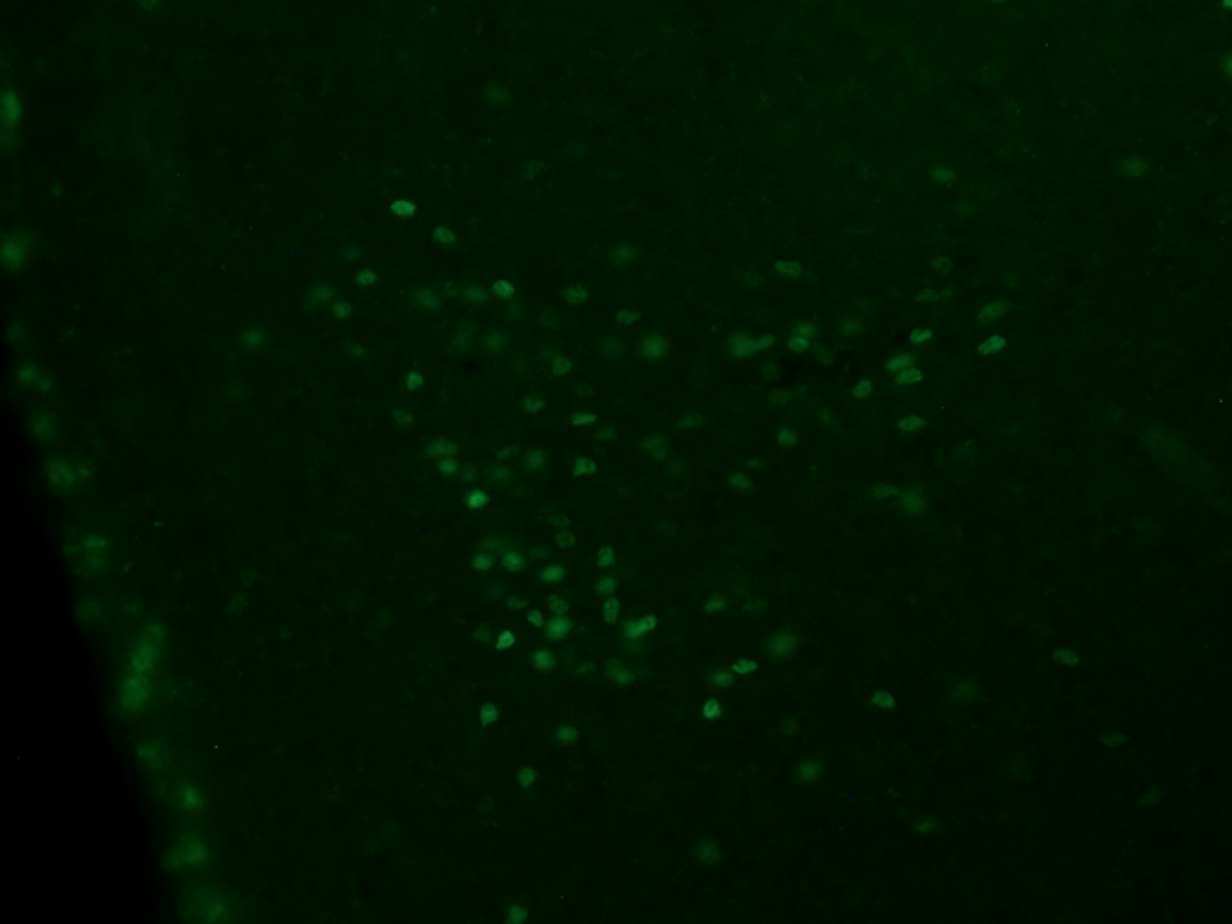}
        \caption{Green image example}
        \label{fig:green_image}
    \end{subfigure}%
    \begin{subfigure}{.33\textwidth}
        \centering
        \includegraphics[width=\linewidth]{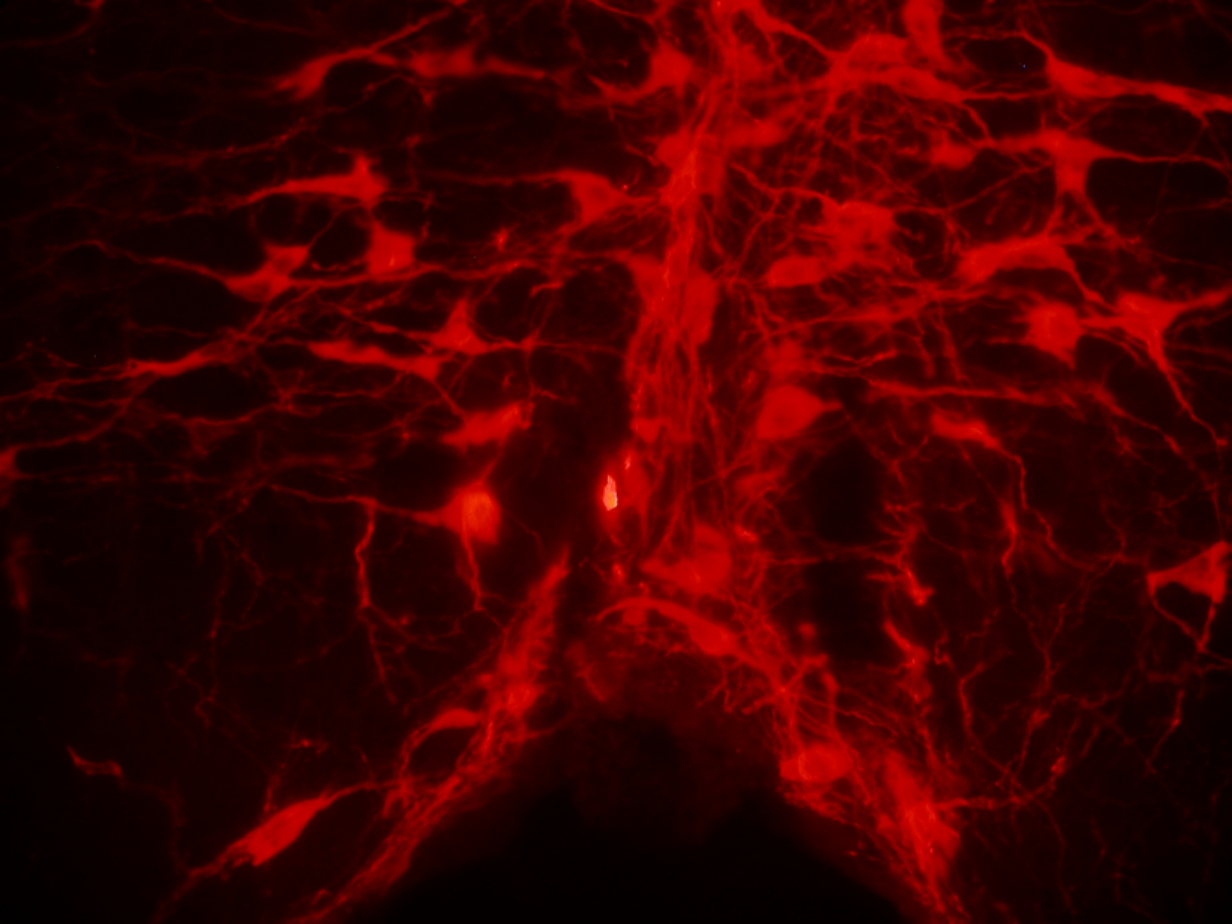}
        \caption{Red image example}
        \label{fig:red_image}
    \end{subfigure}%
    \begin{subfigure}{.33\textwidth}
        \centering
        \includegraphics[width=\linewidth]{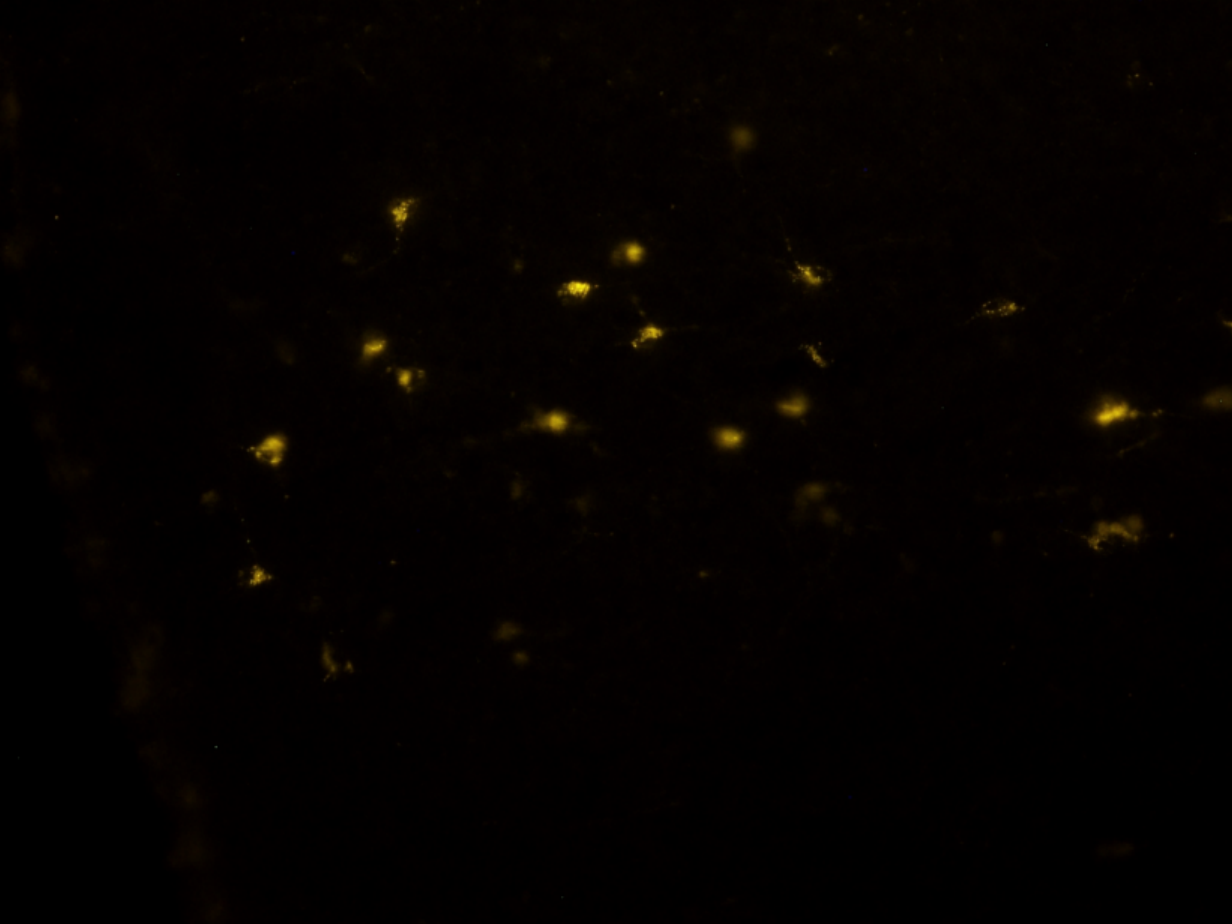}
        \caption{Yellow image example}
        \label{fig:yellow_image}
    \end{subfigure}

    \begin{subfigure}{.33\textwidth}
        \centering
        \includegraphics[width=\linewidth]{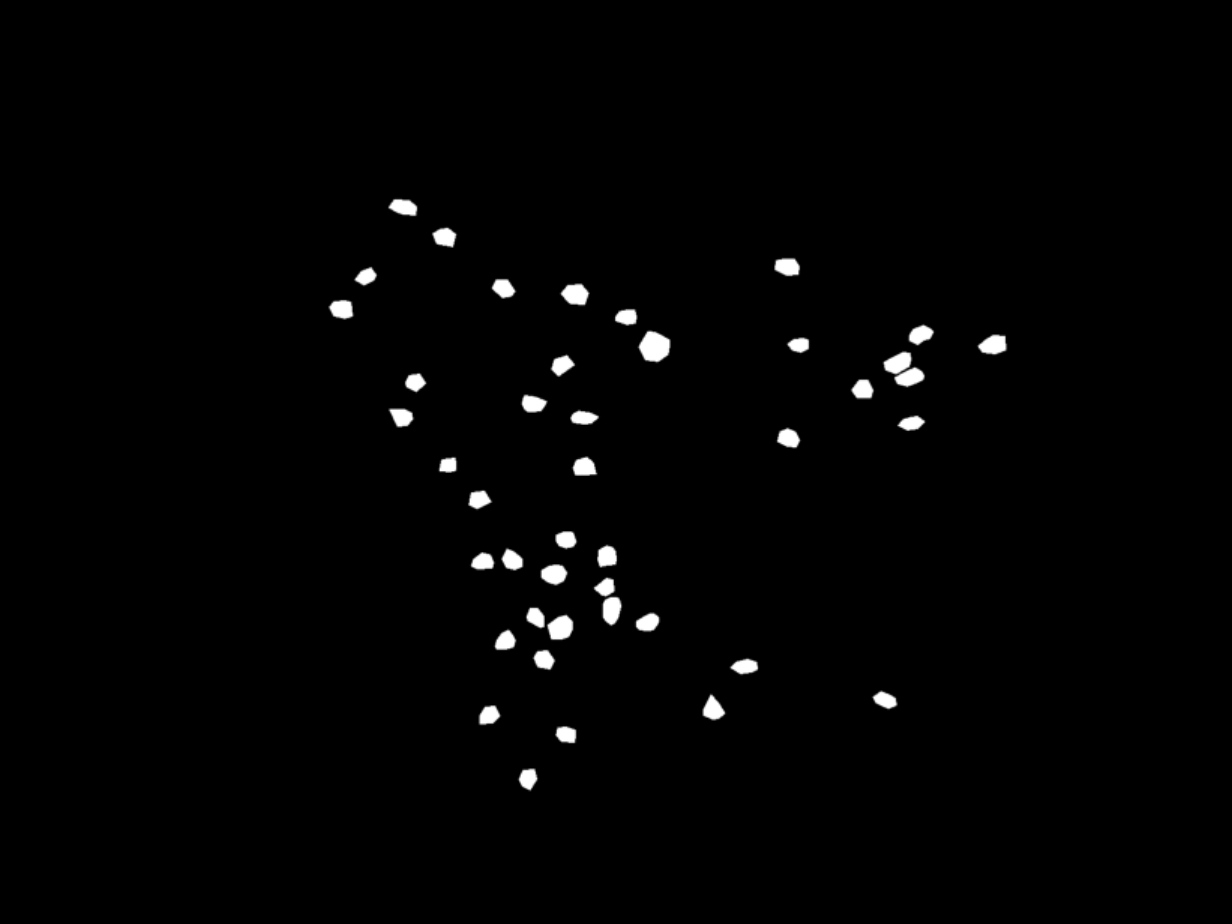}
        \caption{Green mask example}
        \label{fig:green_mask}
    \end{subfigure}%
    \begin{subfigure}{.33\textwidth}
        \centering
        \includegraphics[width=\linewidth]{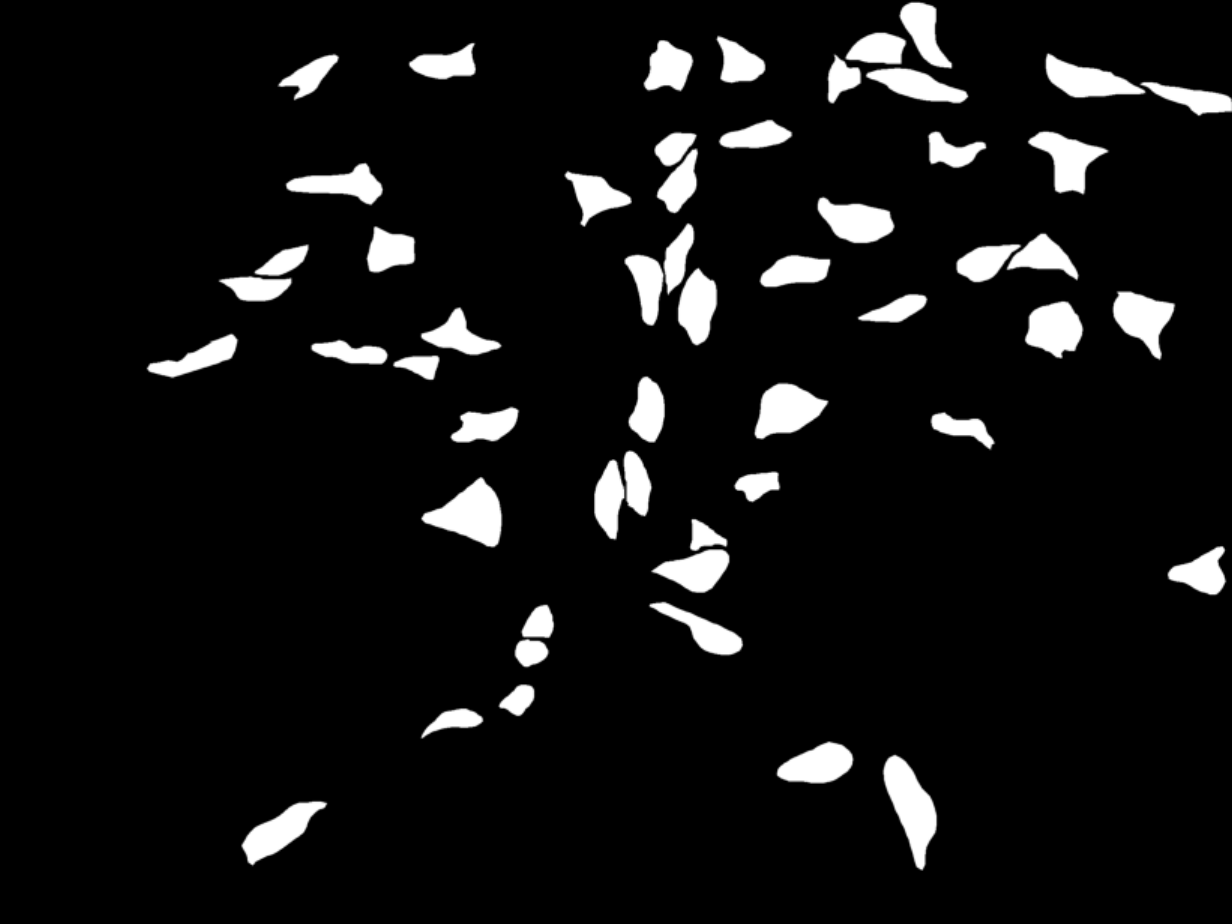}
        \caption{Red mask example}
        \label{fig:red_mask}
    \end{subfigure}%
    \begin{subfigure}{.33\textwidth}
        \centering
        \includegraphics[width=\linewidth]{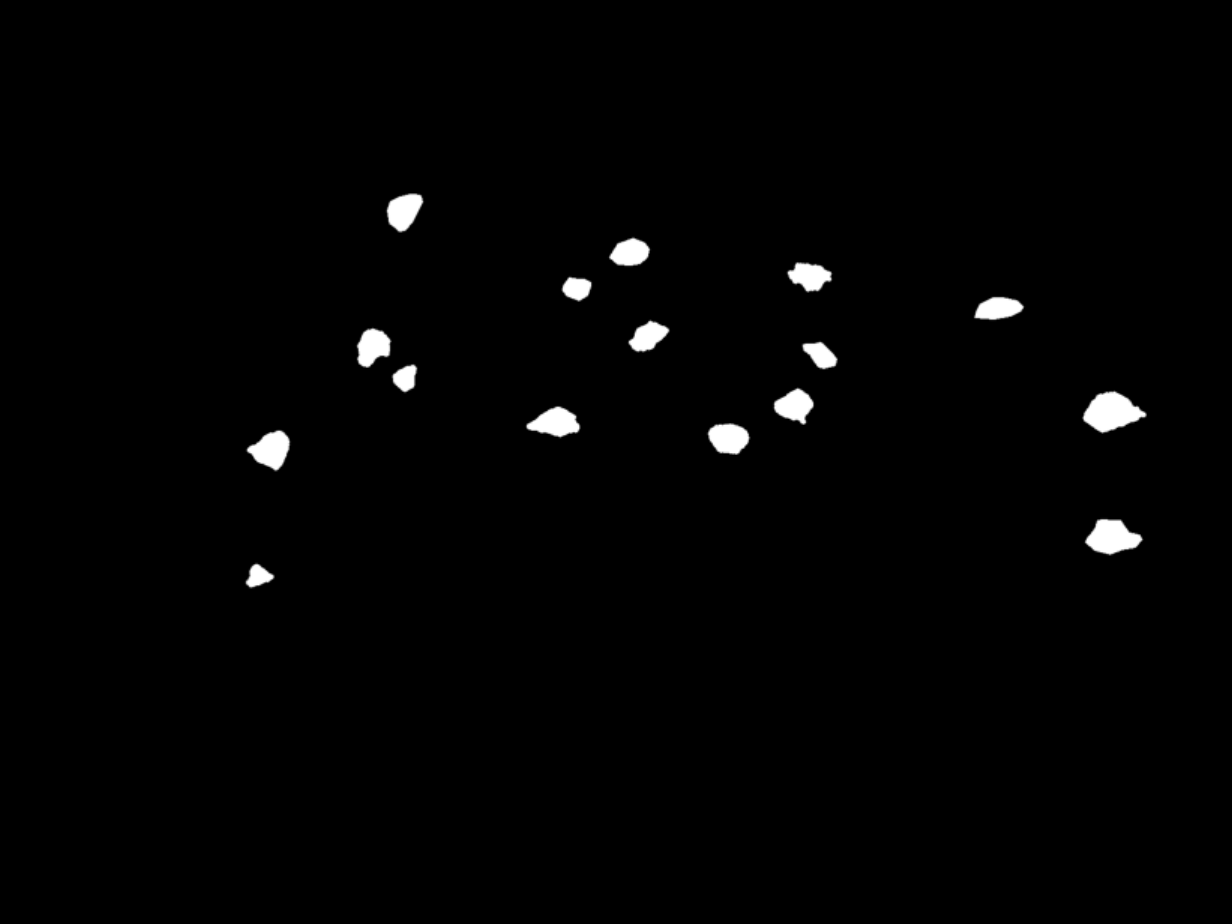}
        \caption{Yellow mask example}
        \label{fig:yellow_mask}
    \end{subfigure}
    
    \caption{\textbf{Data preview}. The figures show examples of fluorescence micrscopy pictures (\Cref{fig:green_image,fig:red_image,fig:yellow_image}) and the corresponding ground-truth binary masks (\Cref{fig:green_mask,fig:red_mask,fig:yellow_mask}).}
    \label{fig:data_preview}
\end{figure}

\subsection*{Image acquisition}

In the image acquisition phase, a total of 68 rodents were subjected to controlled experimental conditions to study torpor and thermoregulatory mechanisms. 
At the end of the experimental session, the animals were deeply anaesthetized and transcardially perfused with 4\% formaldehyde\cite{hitrec2019neural}. 
This process allowed for the tagging of several neuronal substructures located within the nucleus or cytoplasm of the neurons.

Rodents brains were then sectioned into 35 $\mu m$ thick tissue slices, with sampling conducted at regular intervals (105 $\mu m$ for mice and of 210 $\mu m$ for rats) to avoid redundant data and ensure comprehensive coverage while maintaining manageable data size. Brain slices were finally stained for distinct markers following a standard immunofluorescence protocol  \cite{hitrec2019neural}.
Only some areas of interest were observed, namely the Raphe Pallidus (RPa), Dorsomedial Hypothalamus (DMH), Lateral Hypotalamus (LH), and Ventrolateral Periaqueductal Gray (VLPAG).
These specific brain regions were chosen based on their relevance to the study of torpor mechanisms.
The resulting specimens were observed by means of a fluorescence microscope equipped with a high-resolution camera. 
A specific wavelength of excitation light was selected for each collection based on the excitation wavelength of the chosen marker, resulting in pictures acquired with the application of green, yellow/orange or red filters. For simplicity, the image collections are named according to their prevalent hue.
The original images were acquired as either \textit{TIF} or \textit{JPG} files depending on the camera default settings. 
To ensure traceability, a file naming convention was adopted to indicate their respective sample origins\footnote{\texttt{<animal\_id>\_S<sample\_id>C<column\_id>R<row\_id>\_<brain\_area>\_<zoom>\_<collection\_id>}}.
During the analysis phase, the raw data were converted to uncompressed \textit{PNG} format, taking care to preserve the extensive set of associated metadata. 
This conversion aimed to enhance accessibility and facilitate broader utilization of the data, allowing for inspection and manipulation without the need for specialized software.
Consequently, the FNC archive includes both these derived images and the original raw images, which are retained for data recovery and reproducibility purposes.

\subsubsection*{Green and Yellow collections}
The images within these collections were obtained during the same experiment \cite{hitrec2019neural}, in which brain sections from C57BL/6J mice were stained with two markers to highlight specific substructures present in the neurons' nucleus and cytoplasm. The resulting brain slices were then observed using a Nikon Eclipse 80i microscope, equipped with a Nikon Digital Sight DS-Vi1 color camera, at a magnification of 200x.

More specifically, the \textbf{green} collection corresponds to cFOS staining (cf. \Cref{fig:green_mask}). This staining method was employed to emphasize the nuclei of active neuronal cells\cite{kovacs2008measurement}, enabling the topographic analysis of brain areas that exhibit neuronal activity under specific experimental conditions. This approach is widely employed to identify neuronal cells responsible for regulating specific physiological phenomena.

In contrast, the \textbf{yellow} collection (cf. \Cref{fig:yellow_image}) utilized staining for the b-subunit of Cholera Toxin (CTb). This monosynaptic retrograde neuronal tracer migrates within the soma and axons of neuronal cells projecting to the brain area where CTb was previously injected during in vivo experiments\cite{lencer2003intracellular}. Consequently, this staining technique facilitates the identification of morphological connections between different brain regions.

\subsubsection*{Red collection}
The \textbf{red} collection comprises images obtained from multiple unpublished experiments, concerning specimens of both mice and rats (cf. \Cref{fig:red_image}). 
Despite sharing the same experimental setup as green and yellow collections, this time the brain tissues were stained for various elements to phenotypically characterize the cells involved in the neural circuits underlying the physiological phenomena of torpor and thermoregulation.
Specifically, slices were stained for orexin, tryptophan hydroxylase, and tyrosine hydroxylase.
In this case, image acquisition was conducted using both the aforementioned Nikon Eclipse 80i microscope and an ausJENA JENAVAL microscope, equipped with a Nikon Coolpix E4500 color camera, at a magnification of 250x. 
For further details, please refer to the accompanying metadata for each image.

\subsection*{Data annotation}

\begin{figure}[h]
  \centering
  \begin{subfigure}[b]{0.24\textwidth}
    \centering
    \includegraphics[width=\linewidth]{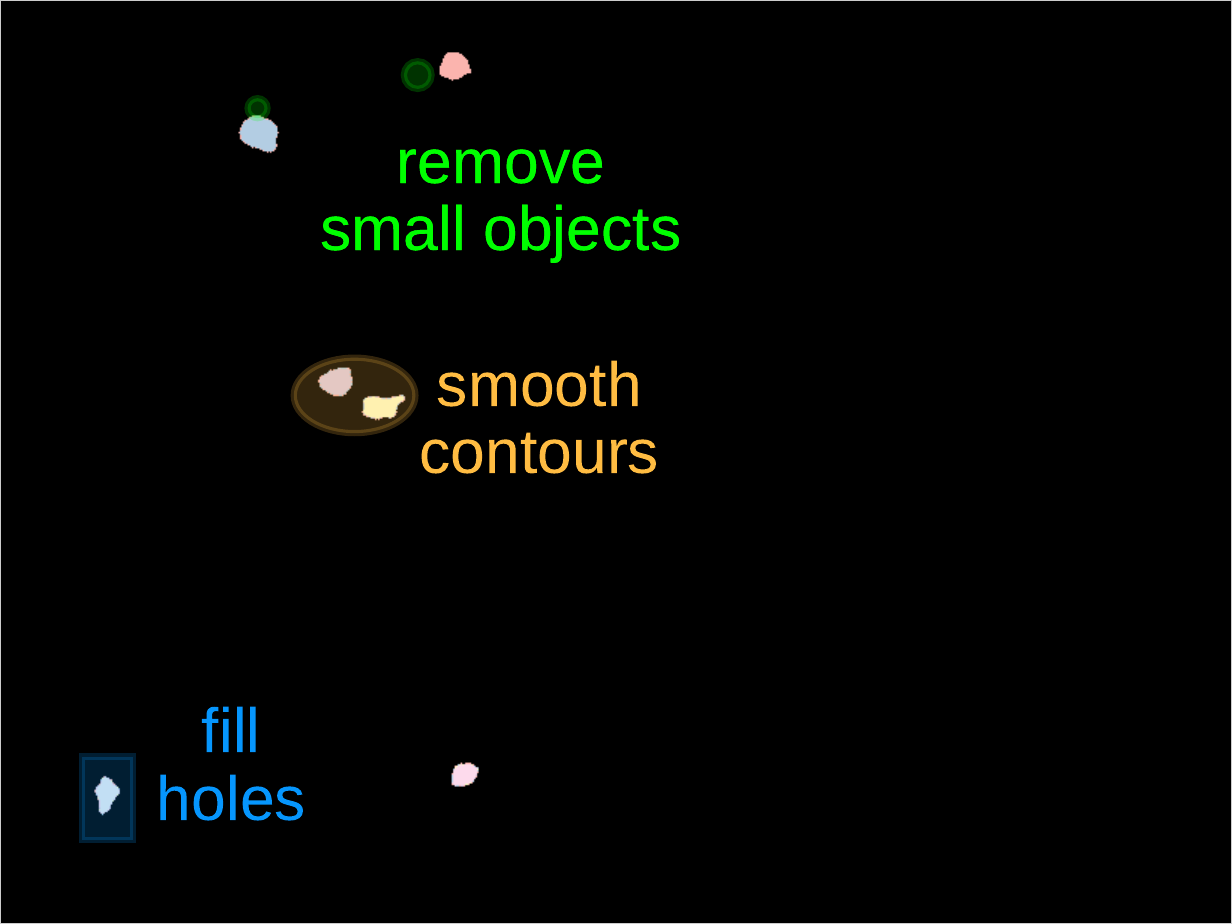} 
    \caption{v2}
    \label{fig:validation1_new}
  \end{subfigure}
  \begin{subfigure}[b]{0.24\textwidth}
    \centering
    \includegraphics[width=\linewidth]{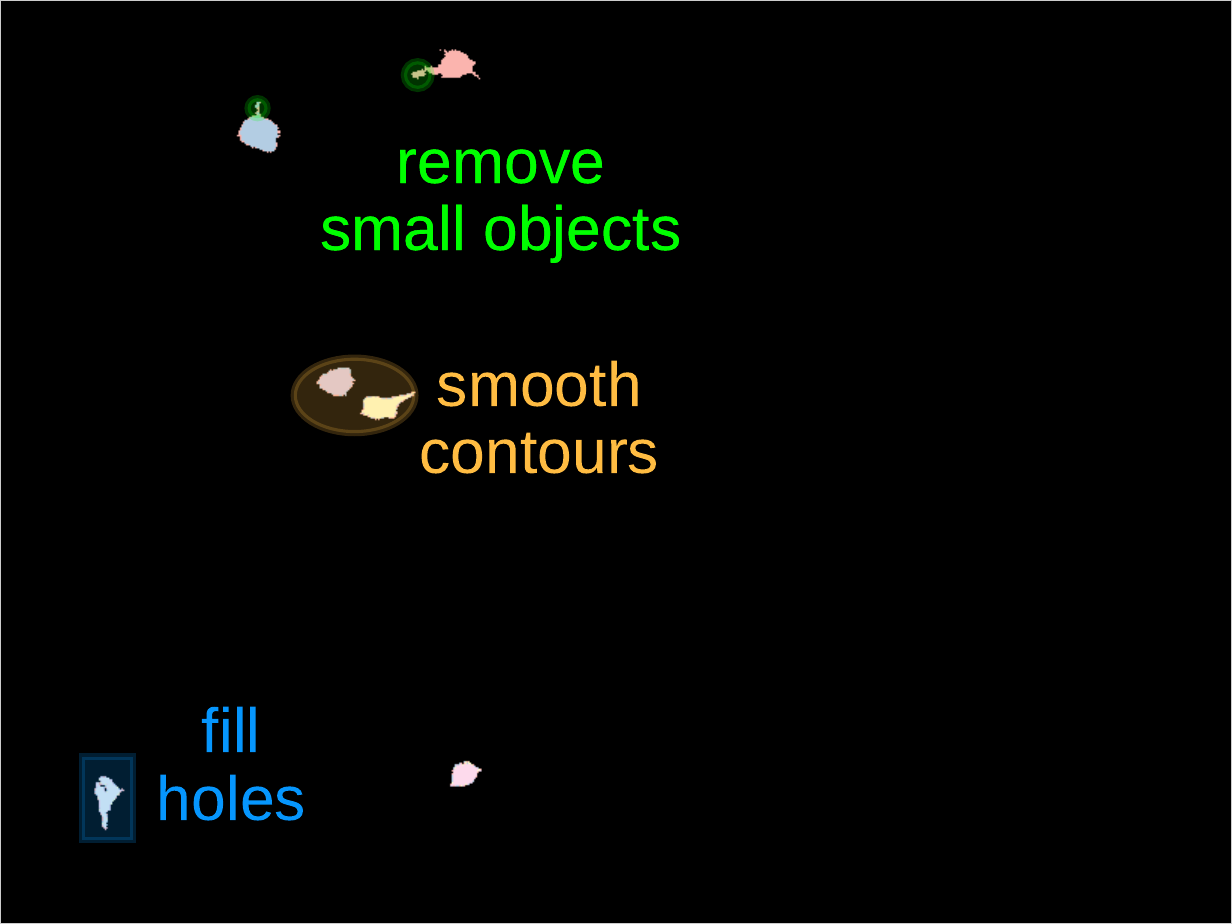}
    \caption{v1}
    \label{fig:validation1_old}
  \end{subfigure}
  \begin{subfigure}[b]{0.24\textwidth}
    \centering
    \includegraphics[width=\linewidth]{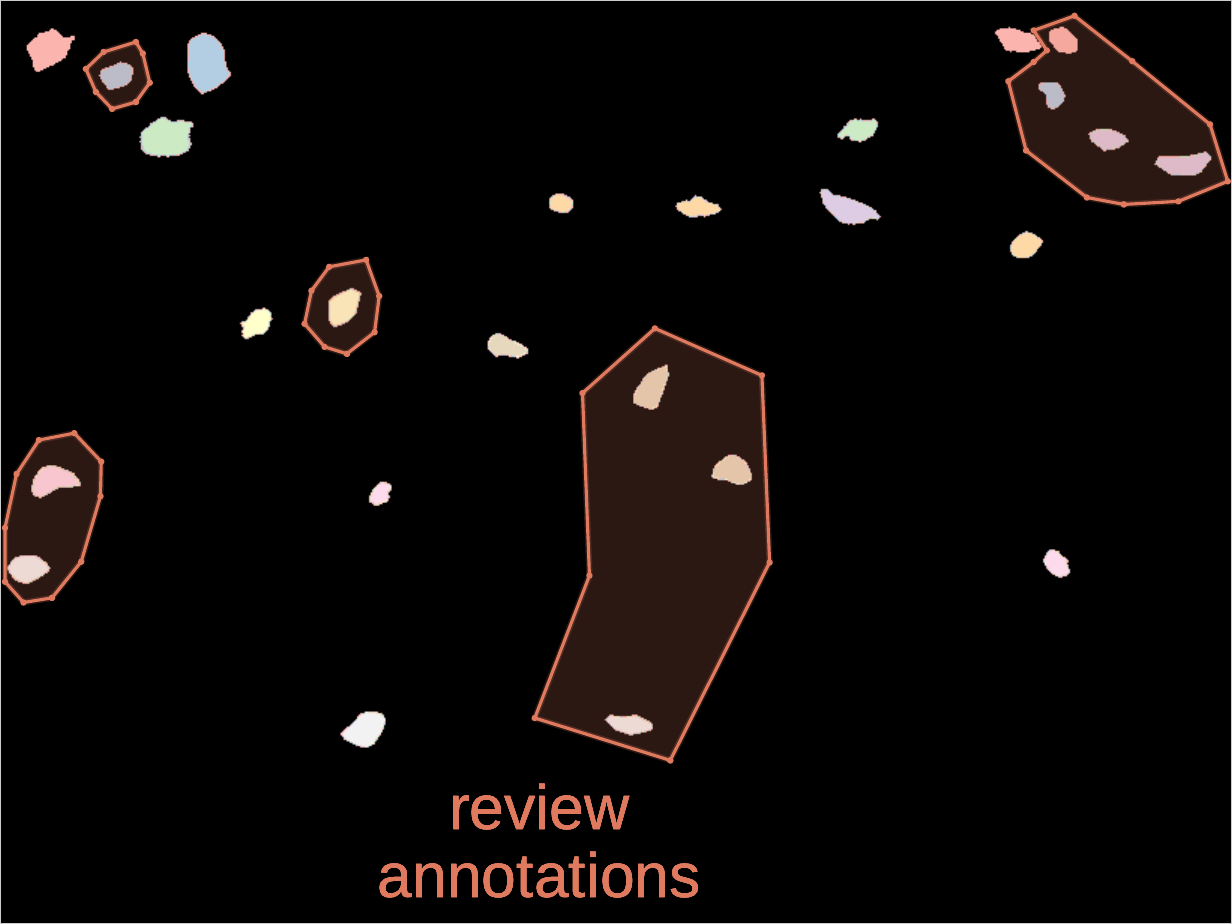}
    \caption{v2}
    \label{fig:validation2_new}
  \end{subfigure}
  \begin{subfigure}[b]{0.24\textwidth}
    \centering
    \includegraphics[width=\linewidth]{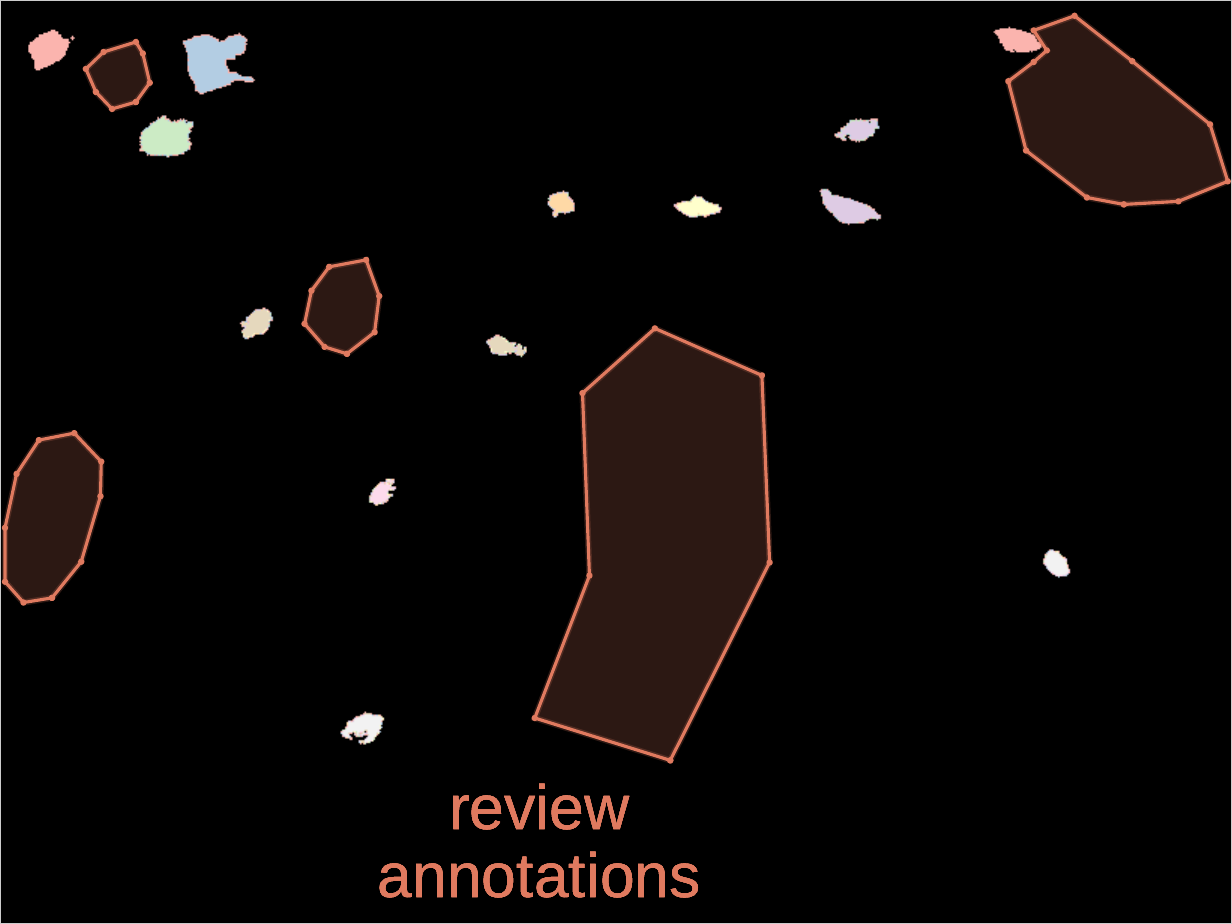}
    \caption{v1}
    \label{fig:validation2_old}
  \end{subfigure}
  \caption{\textbf{Yellow masks v1 review.} \Cref{fig:validation1_new,fig:validation2_new} illustrate how binary masks were reviewed compared to version 1, respectively \Cref{fig:validation1_old,fig:validation2_old}.
  Improvements include: small objects removal, contour smoothing, holes filling and more consistent labelling.}
  \label{fig:yellow_review}
\end{figure}

The data annotation process was carried out by multiple proficient experimenters according to a fixed annotation protocol\footnote{check \textit{Annotations protocol.pdf} inside \textit{Annotations.zip} in the data archive}, with multiple revision rounds to ensure data quality and minimize operator bias.
We adopted the Visual Geometry Group Visual Image Annotator (VIA) annotation tool\cite{dutta2016via,dutta2019vgg}, which employs a web interface for image visualization and allows for the overlaying of annotations in different forms.
In our study, the tagging process involved creating polygon contours, and the resulting annotations were exported into \textit{CSV} format.
To generate the binary masks required for training, the polygon contours were transformed using programming libraries such as OpenCV and scikit-image. 
For the yellow collection, we utilized the binary masks available from version 1\cite{clissa2021fluocells} as pre-annotations. Specifically, we employed erosions and dilations techniques to address fragmented contours resulting from semi-automatic labeling based on thresholding. 
Furthermore, we applied methods to fill small holes within segmented objects, and removed spurious objects that went overlooked in the previous annotations or were erroneously added by prior processing.
Subsequently, these pre-annotations were refined manually using VIA, enhancing their accuracy and ensuring better consistency across the annotations (see \Cref{fig:yellow_review}). In contrast, the green and red collections were annotated from scratch.

Upon completion of the labeling process, the polygon contours exported from VIA were also converted into multiple annotation types and formats. This conversion aims at facilitating accessibility for a wide range of users and promoting the exploration of various learning problems related to our data. For a more comprehensive understanding of the available formats and annotation types, please refer to the \hyperref[sec:data]{Section \textit{Data Records}}.

\section*{Data Records}\label{sec:data}



\begin{figure}
    \centering
    \begin{minipage}[t]{0.25\textwidth}
    \vspace{0pt}
    \vspace{10pt}
    
    \dirtree{%
    .1 \colorbox{Apricot}{\color{white}\textbf{<collection>}}.
    .2 test.
    .3 \colorbox{teal}{\color{white}\textbf{ground\_truths}}.
    .3 images.
    .3 metadata.
    .2 trainval.
    .3 \colorbox{teal}{\color{white}\textbf{ground\_truths}}.
    .3 images.
    .3 metadata.
    .2 unlabelled.
    .3 images.
    .3 metadata.
    }
\end{minipage}
\hspace{0.65pt}
\begin{minipage}[t]{0.25\textwidth}
    \vspace{0pt}
    \vspace{10pt}
    
    \dirtree{%
    .1 \colorbox{teal}{\color{white}\textbf{ground\_truths}}.
    .2 COCO.
    .2 masks.
    .2 Pascal\_VOC.
    .2 rle.
    .2 VIA.
    }
\end{minipage}
    \hspace{0.65pt}
    \begin{minipage}[t]{0.45\textwidth}
        \vspace{0pt}

        \includegraphics[width=\textwidth]{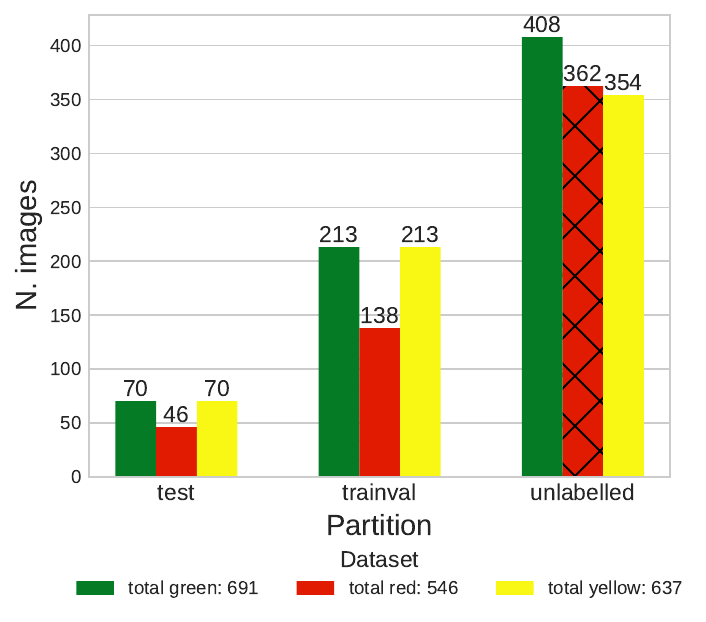}
    \end{minipage}

    \vspace{0.5cm}
    
    \begin{subfigure}[b]{0.25\textwidth}
        \centering
        \caption{Collection folder structure}
        \label{fig:dataset_structure_dirtree}
    \end{subfigure}
    \hspace{0.65pt}
    \begin{subfigure}[b]{0.25\textwidth}
        \centering
        \caption{Annotations folder structure}
        \label{fig:annotations_structure_dirtree}
    \end{subfigure}
    \begin{subfigure}[b]{0.45\textwidth}
        \centering
        \caption{Dataset composition}
        \label{fig:archive-composition}
    \end{subfigure}
    
    \caption{\textbf{FNC dataset structure}.
    \Cref{fig:dataset_structure_dirtree} shows the structure of each image collection folder, while \Cref{fig:annotations_structure_dirtree} gives more details on the organization of the annotations directory. \Cref{fig:archive-composition} summarizes the composition of each image collection, with the amounts of training, testing and unlabelled images.}
    \label{fig:folder_structures_dirtree}
\end{figure}

The FNC dataset is a collection of 1874 high-resolution fluorescent microscopy pictures, 750 of which also have their corresponding ground-truth segmentation masks, while the remaining 1124 are unlabelled.
It is hosted on \href{https://amsacta.unibo.it/id/eprint/7347}{AMS Acta}\footnote{available at: \href{https://doi.org/10.6092/unibo/amsacta/7347}{https://doi.org/10.6092/unibo/amsacta/7347} (link will be active after acceptance)}, the open access repository managed by the University of Bologna.
The data are organized into three standalone image collections, named for simplicity \textit{green}, \textit{yellow}, and \textit{red}, each available under the corresponding folder (see \Cref{fig:archive-composition}). 
The collections share a common layout to facilitate easy access and analysis (see \Cref{fig:dataset_structure_dirtree}).

To aid users in navigating the archive, the \texttt{metadata\_v2.xlsx} file provides a comprehensive overview of the FNC data collection.
It includes high-level metadata for each image, such as the corresponding animal, acquisition details, data partition, and annotation information.

\subsection*{Image collection folder structure}

The \texttt{trainval} and \texttt{test} folders contain all labelled images for each collection. These data partitions were obtained through a random 75\%/25\% split and are recommended as a suggested configuration to ensure reproducibility and comparability in future studies.
The remaining images were collected under the \texttt{unlabelled} folder.

Inside each data partition folder, the \texttt{images} folder contains fluorescence microscopy images in \textit{PNG} format.
All the images are accompanied by a rich set of metadata, stored both in their \textit{EXIF} tags and as a separate \textit{TXT} file under the \texttt{metadata} folder.
The \texttt{ground\_truths} folder contains annotations in various formats commonly used within the machine learning community (see \Cref{fig:annotations_structure_dirtree}).

\subsection*{Annotation types and formats}

The FNC collection provides annotations of multiple \textit{types}, encoded in several standard \textit{formats}.
In the \texttt{masks} folder, we find the binary masks typically used for segmentation tasks (cf. \Cref{fig:green_mask,fig:red_mask,fig:yellow_mask}). The correspondence between the masks and the respective images can be established based on the filenames.
The other folders store a light-weight encoding of the binary masks, enriched with additional annotation types/formats.

The \texttt{rle} directory contains \textit{Running Length Encoding (RLE)} of the binary masks, stored as \textit{pickle} files. 
This encoding is a compressed representation that can effectively save disk space while preserving the complete segmentation information. It is particularly convenient for high-resolution images like those present in our dataset.

The other directories provide several annotation types, and they are named after their annotation format.
\textbf{Polygon} annotations are available in each of the \texttt{VIA}, \texttt{COCO} and \texttt{Pascal\_VOC} directories, in the form of \textit{json} or \textit{xml} files. 
\texttt{COCO}\cite{COCO} and \texttt{Pascal\_VOC}\cite{everingham2010pascal} formats also features \textbf{bounding boxes} and \textbf{dot} annotations for object detection tasks, and \textbf{count} labels for object counting.

\section*{Technical Validation}


In order to demonstrate the potential for successful model training and analysis using the provided dataset, we conducted three types of checks to ensure the accuracy and quality of the annotations.

Firstly, polygon annotations were obtained by experienced researchers. During this phase, the annotations underwent multiple rounds of double-checking to ensure that the polygons did not have intersecting edges and that they accurately represented the objects when transformed into binary masks.

Secondly, we leveraged domain knowledge to validate the annotations. 
Precisely, we tested the binary masks against our expectations regarding the sizes and shapes of the biological structures involved. This validation process relies on a quantitative evaluation concerning objects' area and diameter, complemented by a visual scrutiny of the masks to ensure they align with the expected shapes and exhibit smooth contours.
\begin{table}[htbp]
\centering
\resizebox{\textwidth}{!}{
\begin{threeparttable}
\begin{tabular}{l *{3}{S[table-format=2.2]} c *{3}{S[table-format=2.0]} c *{3}{S[table-format=3.2]} c *{3}{S[table-format=2.2]} c *{3}{S[table-format=2.2]}}
\toprule
& \multicolumn{3}{c}{\textbf{signal \%}} && \multicolumn{3}{c}{\textbf{cell count}} && \multicolumn{3}{c}{\textbf{area (\si{\micro\meter\squared})}} && \multicolumn{3}{c}{\textbf{Feret diameter (\si{\micro\meter})}} && \multicolumn{3}{c}{\textbf{equivalent diameter (\si{\micro\meter})}} \\
\cmidrule(lr){2-4} \cmidrule(lr){6-8} \cmidrule(lr){10-12} \cmidrule(lr){14-16} \cmidrule(l){18-20}
\textbf{collection} & {green} & {red} & {yellow} && {green} & {red} & {yellow} && {green} & {red} & {yellow} && {green} & {red} & {yellow} && {green} & {red} & {yellow} \\ \midrule
count & \multicolumn{1}{c}{283} & \multicolumn{1}{c}{184} & \multicolumn{1}{c}{283} && \multicolumn{1}{c}{4606}\tnote{a} & \multicolumn{1}{c}{4486}\tnote{a} & \multicolumn{1}{c}{2659}\tnote{a} && \multicolumn{1}{c}{4600} & \multicolumn{1}{c}{4483} & \multicolumn{1}{c}{2621} && \multicolumn{1}{c}{4600} & \multicolumn{1}{c}{4483} & \multicolumn{1}{c}{2621} && \multicolumn{1}{c}{4600} & \multicolumn{1}{c}{4483} & \multicolumn{1}{c}{2621} \\
mean  & 0.64 & 2.61 & 0.65 && 28.65 & 34.96 & 25.76 && 74.78 & 246.52 & 132.55 && 12.20 & 25.68 & 17.24 && 9.64 & 17.13 & 12.63 \\
std   & 0.57 & 1.74 & 0.80 && 17.49 & 17.90 & 19.88 && 23.39 & 130.24 & 66.02 && 1.92 & 8.86 & 4.86 && 1.48 & 4.50 & 3.04 \\
\midrule
min   & 0 &   0 &   0 &&    0 &    0 &    0 &&   12.41 &   24.81 &   17.62 && 5.48 &    7.08 &    6.09 && 3.97 &    5.62 &    4.74 \\
10\%   &     0.10 &   0.37 &   0 &&    9 &   14 &    4 &&   48.74 &  106.72 &   66.16 && 9.85 &   15.67 &   11.93 &&                7.88 &   11.66 &    9.18 \\
25\%   &     0.25 &   1.34 &   0.13 &&   15 &   21 &    9 &&   58.58 &  144.04 &   88.02 &&              10.88 &   18.64 &   13.90 &&                8.64 &   13.54 &   10.59 \\
50\%   &     0.45 &   2.34 &   0.40 &&   25 &   31 &   19 &&   71.67 &  218.80 &  117.75 &&              12.07 &   24.38 &   16.51 &&                9.55 &   16.69 &   12.24 \\
75\%   &     0.92 &   3.53 &   0.82 &&   39.75 &   50 &   44 &&   87.33 &  320.75 &  161.17 &&              13.35 &   30.99 &   19.76 &&               10.54 &   20.21 &   14.33 \\
90\%   &     1.38 &   4.89 &   1.49 &&   56 &   58 &   57 &&  104.66 &  422.64 &  218.27 &&              14.70 &   37.64 &   23.59 &&               11.54 &   23.20 &   16.67 \\
max   &     3.36 &   8.40 &   5.54 &&   69 &   90 &   72 &&  227.04 &  842.66 &  548.20 &&              21.21 &   74.39 &   41.66 &&               17 &   32.76 &   26.42 \\
\bottomrule
\end{tabular}
\begin{tablenotes}
\item[a] The difference compared to counts in following columns comes from the contribution of empty images. These amount to 6, 3 and 38 images for green, red and yellow collections, respectively.
\end{tablenotes}
\end{threeparttable}
}
\caption{Summary statistics of key features' distribution for each image collection. The top portion highlights global indicators, while the bottom one reports given percentiles of each distribution.} \label{tab:summary-stats}
\end{table}
\Cref{tab:summary-stats} reports summary statistics for the distribution of key features at the image and object levels, that can be leveraged for technical validation. 
For instance, the annotated objects display an average area of nearly 75, 247, and 133 $\mu m^2$ for green, red, and yellow cells, respectively. These values align with the expected size of the biological structures represented in each image collection.
Additionally, the analysis of Feret and equivalent diameters provides an understanding of the typical form of the stained objects. In particular, the Feret diameter\cite{pabst2007feret} can be interpreted as a measure of the maximum extension of an object, whereas the equivalent diameter represents the diameter the object would have if it were a perfect circle with the same area. Thus, comparing these two metrics can offer insight into the objects' shape regularity.
For green cells, the values for the two measurements are relatively close (roughly 12 VS 10 $\mu m$), suggesting that these cells are broadly circular or oval in shape.
A similar conclusion can be drawn for the yellow stains, albeit with slightly more variability (approximately 17 and 13 $\mu m$), indicating generally regular shapes with occasional deviations.
In the case of red objects, instead, the comparison is markedly different. This time we observe a Feret diameter around 26 $\mu m$ against an equivalent diameter of 17, which suggests that these stains are typically elongated in one direction rather than concentrated around a center of mass.
All these observations are also corroborated when visually inspecting annotated cells, which confirms prior expectations about objects size and shapes based on the nature of the marked structures.

\subsection*{Learning}
Thirdly, we conducted a sample training phase for each image collection using a \textbf{cell ResUnet} architecture\cite{morelli2021cresunet}, specifically designed for this type of application. 
Specifically, we trained a network from scratch for each collection using a Dice Loss\cite{sudre2017diceloss} and the Adam optimizer\cite{kingma2014adam}. The initial learning rate was set based on the ``learning rate test"\cite{smith2019hyperparms} implemented by \textit{fastai}'s\cite{howard2020fastai} \texttt{lr\_find()} method. The training phase continued for 100 epochs with cyclical learning rates\cite{smith2017cyclical,smith2019super}, and the best model was selected based on the best validation dice coefficient. For all technical details please refer to the GitHub repository\footnote{available at: \href{https://github.com/clissa/fluocells-scientific-data}{https://github.com/clissa/fluocells-scientific-data} (link will be active after acceptance)}.
This training phase aims to verify the effectiveness of the data in facilitating the learning of beneficial cell features. Additionally, the intent is to highlight the relevant metrics for result evaluations. 
In particular, we suggest performance should be assessed differently depending on the end goal of future analyses.

For \textbf{segmentation} tasks, we provided an implementation where matching of actual and predicted neurons\footnote{by this we intend the calculation of True Positives (TP), False Positives (FP) and False Negatives (FN)} is done based on their overlap, quantified as \textit{Intersection-over-Union (IoU)}\cite{kirillov2019panoptic}. This approach not only ensures a 1-1 correspondence of true and predicted objects but also assesses how closely the predictions reconstruct the shape of ground-truth cells.
Building on top of this definition, standard metrics such as \textit{precision, recall} and \textit{$F_1$ score} can be computed as measures of global performance.

\textbf{Detection} tasks, on the other hand, would benefit from a looser matching criterion, comparing predicted and true objects' centers instead of overlaps. This approach prioritizes recognition over precise shape reconstruction.

Finally, for \textbf{counting} tasks we suggest common regression metrics such as \textit{Mean Absolute Error (MAE), Median Absolute Error (MedAE)} and \textit{Mean Percentage Error (MPE)}. 

\begin{table}[ht]
\centering
\label{tab:performance-metrics}
\begin{tabular}{@{}lcccccccccc@{}}
\toprule
& \multicolumn{3}{c}{\textbf{segmentation}} & \multicolumn{3}{c}{\textbf{detection}} & \multicolumn{3}{c}{\textbf{counting}} \\
\cmidrule(lr){2-4} \cmidrule(lr){5-7} \cmidrule(lr){8-10}
\textbf{metrics} & $F_1$ Score & Precision & Recall & $F_1$ Score & Precision & Recall & MAE & MedAE & MPE (\%) \\ \midrule
Green & 0.69 & 0.79 & 0.62 & 0.77 & 0.86 & 0.69 & 1.20 & 1.00 & 9\% \\
Red & 0.28 & 0.33 & 0.25 & 0.60 & 0.69 & 0.53 & 7.13 & 7.00 & 42\% \\
Yellow & 0.65 & 0.67 & 0.63 & 0.77 & 0.78 & 0.75 & 1.54 & 1.00 & 18\% \\ \bottomrule
\end{tabular}
\caption{\textbf{Performance metrics by learning task.} The \textit{segmentation} portion refers to TP, FP, and FN computed based on objects overlapping (IoU). For \textit{detection} metrics, predicted and true cells are associated based on their centers' distance. Counting metrics simply consider the difference between predicted and true objects.}\label{tab:metrics}
\end{table}

\Cref{tab:metrics} shows the results of the sample training for each image collection. These results are not intended to be a comprehensive exploration of the model's capabilities on FNC data, but rather to showcase some characteristics of various evaluation methods.  Nonetheless, they may serve as a baseline for future studies.

Despite no optimization of the training pipeline, the initial results are mainly satisfactory (except for red segmentation), confirming the technical robustness of the data collection process.
Going into more details, we observe a marked discrepancy between segmentation and detection metrics. As expected, the $F_1$ scores based on the distance between true and predicted centers of mass are significantly higher than the corresponding segmentation indicators. 
Moreover, the discrepancy is greater for image collections where the objects have more irregular shapes (green < yellow < red).
This is a consequence of the more inclusive matching criterion used for detection tasks.

In terms of counting, performance is already very satisfactory. However, these metrics may not fully represent the model's performance as good results could arise due to a balancing effect between true positives and false negatives. 
Interestingly, despite low absolute errors, the percentage error is relatively high due to the impact of errors in images with few or no cells. 
To address this issue, we adopt the following formula for MPE computation: 
$ \text{MPE} = \dfrac{\left( \text{predicted} - \text{true} \right)}{\max{\left( \text{true}, 1\right)}}$.
In this way, the fraction is not over-inflated when there are no cells in the original mask.

\section*{Usage Notes}\label{sec:usage}



The Fluorescent Neuronal Cells collection is available both as a comprehensive archive and as individual image collections for specific research requirements. This enables users to download the data efficiently and selectively, based on their specific needs.
The code provided is based on the Python and PyTorch frameworks, offering a robust foundation for analysis and modeling. However, thanks to the popularity of the annotation formats and the use of PNG images, users can easily employ their preferred deep learning framework.

\subsection*{Peculiar traits}

In all image collections, the visual representation is characterized by the prevalence of two distinct color tones, which result from the deliberate selection of a specific wavelength. One tone appears darker, indicating areas where light has been filtered out, while the other tone is brighter and more intense, emitted by the fluorophore corresponding to the color of each collection (see \Cref{fig:green_image,fig:red_image,fig:yellow_image}).
As a result, the images can generally be depicted using variations of a single color. Consequently, a 1-D representation may be sufficient, or an alternative color space other than RGB could provide more informative and less redundant data.

Notice, however, that the specific colors employed in our studies were dictated not by any inherent or functional property of the stained biological structures, but rather by their accessibility and practicality during the time of the experiments. Therefore, it would be a misinterpretation to associate specific colors to particular neuronal substructures. In fact, these colors serve only as contrasting elements to discern the stained foreground objects from the background. 
Consequently, the emphasis should lie primarily on learning this discrimination rather than matching specific colors with the neuronal structures. 
Thus, the particular colors should not be considered indicative of the type of neuronal cells or their functional attributes, but merely as a practical tool aiding in the overall visualization and interpretation.

\begin{figure}[htbp]
    \centering
    \begin{subfigure}{0.3\textwidth}
        \includegraphics[width=\textwidth]{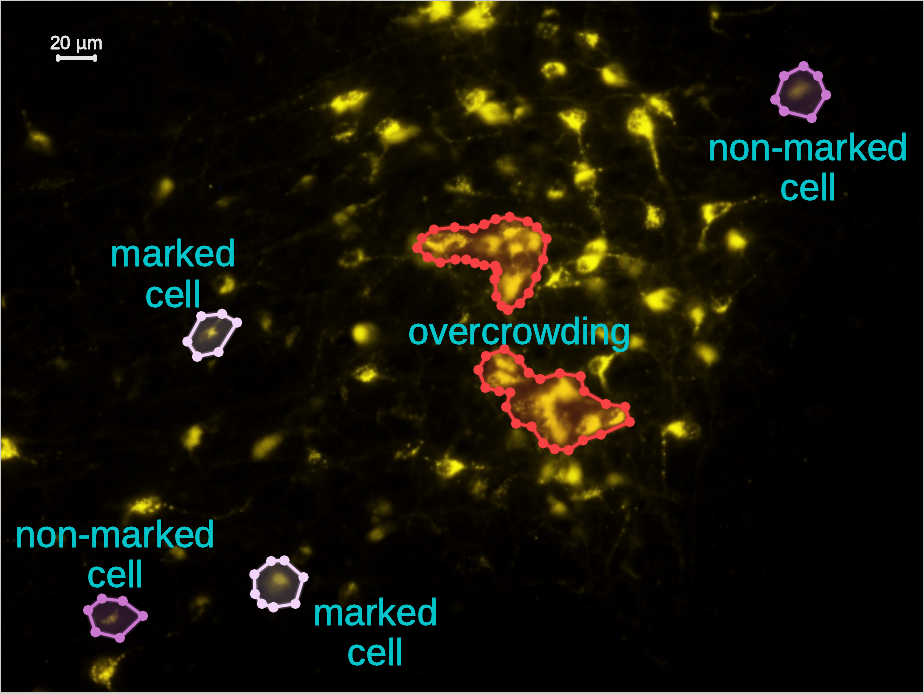}
        \caption{}\label{fig:challenges:yellow}
    \end{subfigure}
    \begin{subfigure}{0.3\textwidth}
        \includegraphics[width=\textwidth]{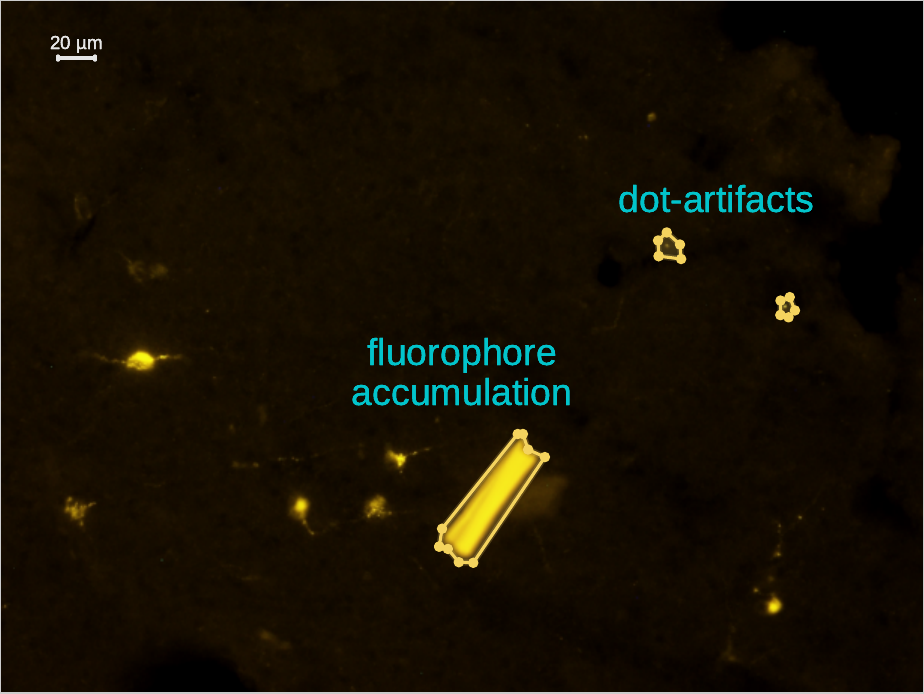}
        \caption{}\label{fig:challenges:yellow_artifact}
    \end{subfigure}
    \begin{subfigure}{0.3\textwidth}
        \includegraphics[width=\textwidth]{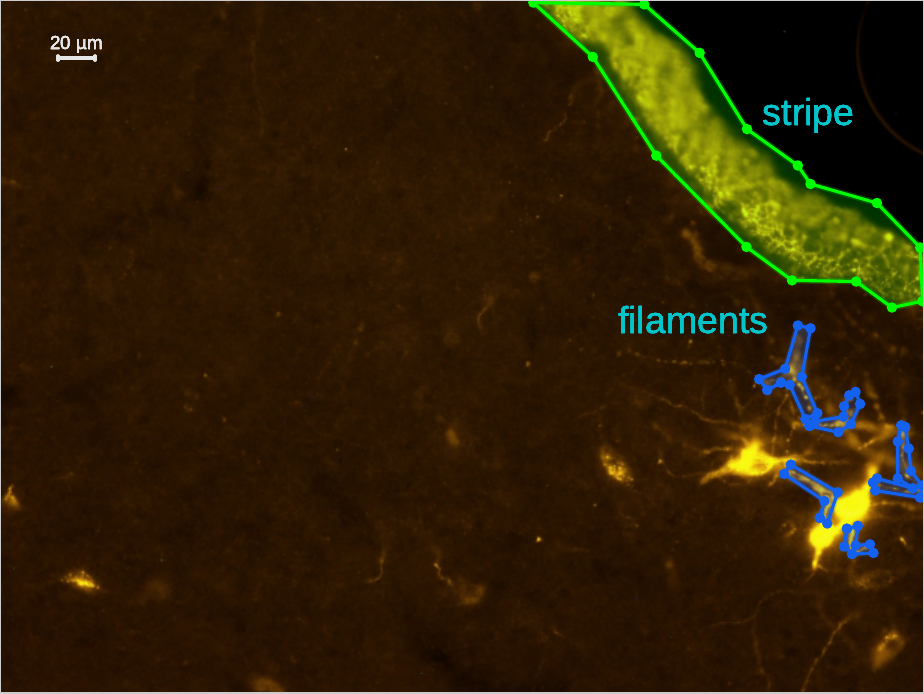}
        \caption{}\label{fig:challenges:yellow_stripe}
    \end{subfigure}
    
    \vspace{1em} 

    \begin{subfigure}{0.23\textwidth}
        \includegraphics[width=\textwidth]{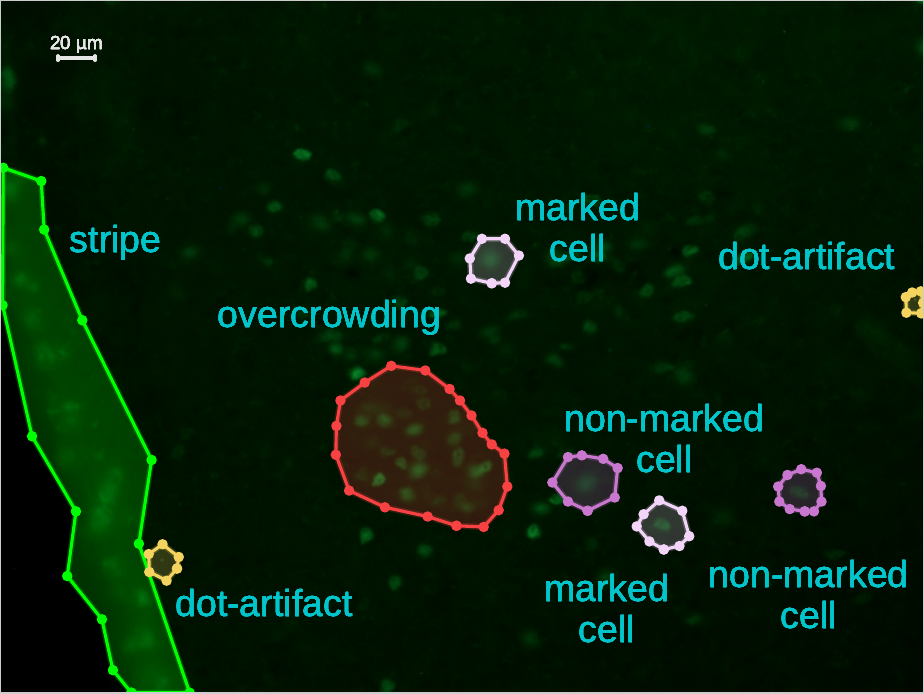}
        \caption{}\label{fig:challenges:green}
    \end{subfigure}
    \begin{subfigure}{0.23\textwidth}
        \includegraphics[width=\textwidth]{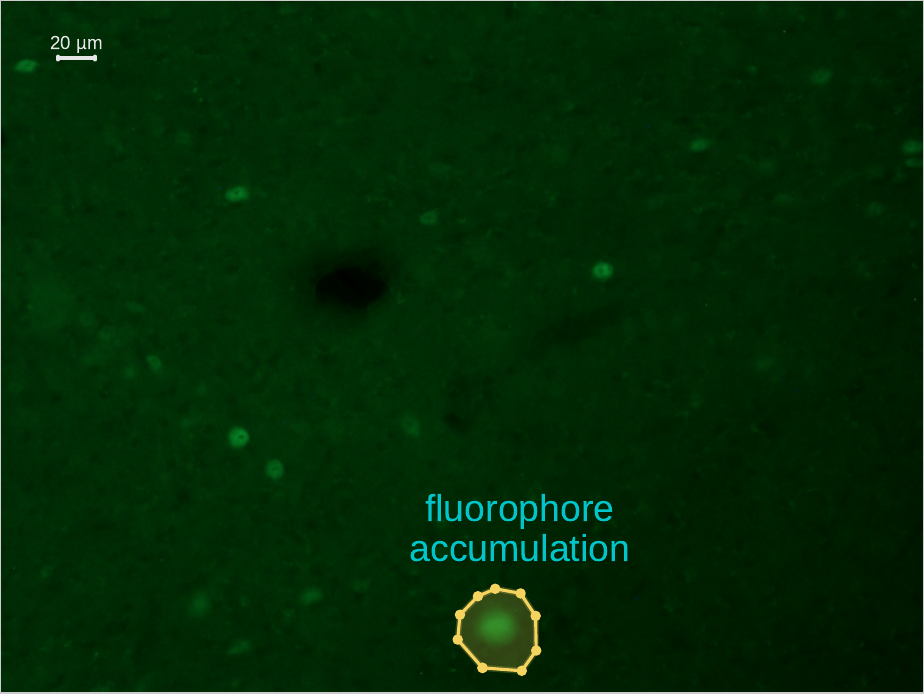}
        \caption{}\label{fig:challenges:green_artifact}
    \end{subfigure}
    \begin{subfigure}{0.23\textwidth}
        \includegraphics[width=\textwidth]{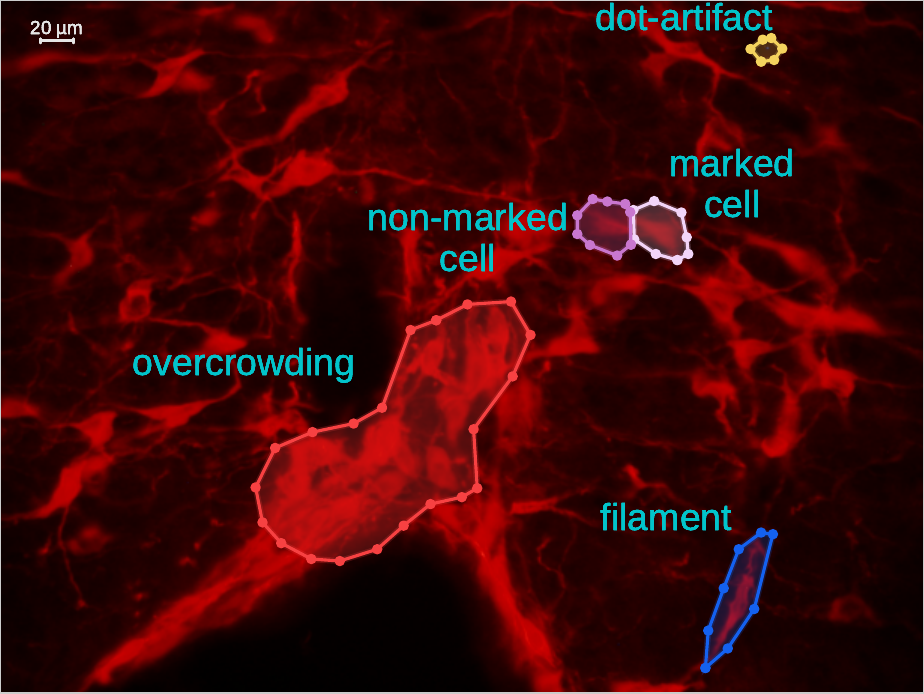}
        \caption{}\label{fig:challenges:red_filament}
    \end{subfigure}
    \begin{subfigure}{0.23\textwidth}
        \includegraphics[width=\textwidth]{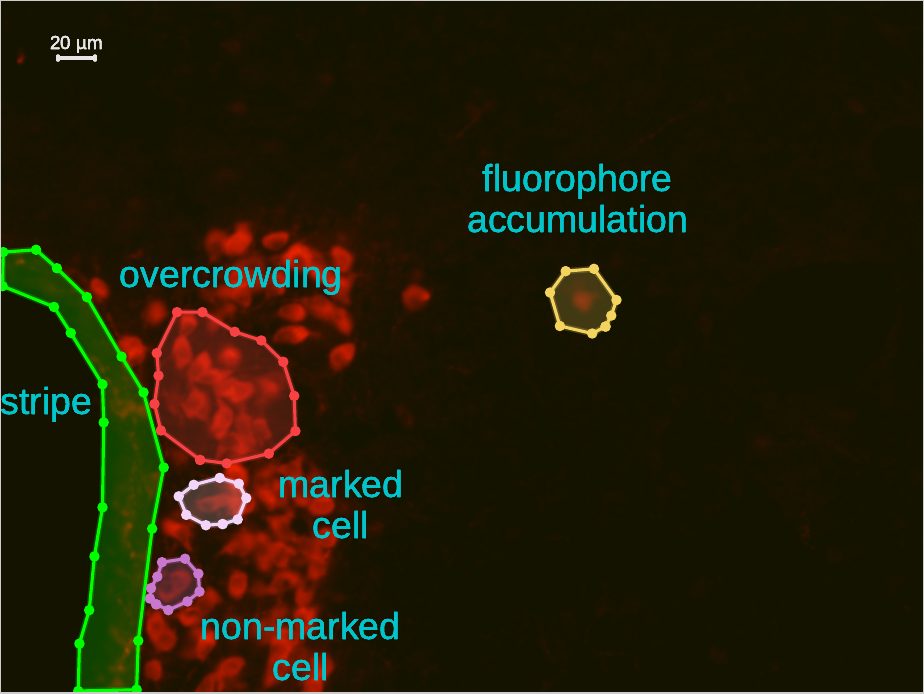}
        \caption{}\label{fig:challenges:red_stripe}
    \end{subfigure}

    \caption{\textbf{Challenges}. FNC data present several difficulties to take into account during their analysis. Common challenges are represented by \textit{overcrowding} (\Cref{fig:challenges:yellow,fig:challenges:green,fig:challenges:red_filament,fig:challenges:red_stripe}), \textit{ambiguity} (\Cref{fig:challenges:yellow,fig:challenges:green,fig:challenges:red_filament}), and \textit{artifacts} (\Cref{fig:challenges:yellow_artifact,fig:challenges:yellow_stripe,fig:challenges:green,fig:challenges:green_artifact,fig:challenges:red_filament,fig:challenges:red_stripe}).}\label{fig:challenges}
\end{figure}

\subsection*{Challenges}

Some important insights for future studies can be drawn examining ground-truth masks at the pixel level, revealing significant characteristics that impact the training process.

The two classes, namely cells (1) and background (0), exhibit an extreme \textbf{class imbalance}, with background pixels being overwhelmingly predominant, typically exceeding cell pixels by over a factor of 100 (cf. \Cref{tab:summary-stats}, \textit{signal \%}).
These observations highlight the necessity for specialized training strategies to address this pronounced class imbalance and effectively learn the pixel classification.

Additional challenges are associated with the macroscopic content of the images. The Fluorescent Neuronal Cells data showcase a diverse collection of 11704 subnuclear neuronal structures, varying in shape, size, and extension (cf \Cref{tab:summary-stats}, \textit{area, Feret diameter} and \textit{equivalent diameter} columns). 
The distribution of these structures across the collections is uneven, with some images containing numerous cells while others are devoid of them. Consequently, the model needs to be flexible enough to handle both scenarios.

Furthermore, despite considerable efforts to stabilize the acquisition procedure, several technical challenges persist.
Firstly, there is a \textbf{high variability in terms of color, saturation, and contrast} from one image to another. For instance, there are instances where the tissues absorb some of the markers (see \Cref{fig:challenges:yellow_artifact,fig:challenges:yellow_stripe,fig:challenges:green,fig:challenges:green_artifact,fig:challenges:red_stripe}), causing irrelevant compounds to emit light which is then captured by the microscope. 
Consequently, the background's hue may shift towards values similar to those of faint neuronal cells (see \Cref{fig:challenges:yellow_artifact,fig:challenges:yellow_stripe,fig:challenges:green,fig:challenges:green_artifact,fig:challenges:red_filament}).
In such circumstances, relying solely on pixel intensity is insufficient to distinguish between signal and background, necessitating the consideration of additional characteristics such as saturation and contrast. However, even the analysis of these characteristics is not straightforward, as fluorescent emissions are naturally unstable, leading to fluctuations in the saturation levels exhibited by cell pixels (cf. \Cref{fig:challenges:yellow,fig:challenges:yellow_artifact,fig:challenges:yellow_stripe} or \Cref{fig:challenges:red_filament,fig:challenges:red_stripe}).

Moreover, the substructures of interest have a fluid nature. Also, the shot can capture different two-dimensional sections depending on how the cells are oriented within the tissues.
As a consequence, the \textbf{size and the shape of the stained cells can vary significantly} (cf. objects dimension in \Cref{fig:green_mask,fig:red_mask,fig:yellow_mask}), further complicating the discrimination between cells and the background.

Another challenge arises from the occasional presence of accumulations of fluorophore in narrow areas, resulting in emissions that closely resemble those of cells. 
These \textbf{artifacts} can manifest as small areas, such as point artifacts and filaments, or larger structures, like lateral stripes  (see \Cref{fig:challenges:yellow_artifact,fig:challenges:yellow_stripe,fig:challenges:green,fig:challenges:green_artifact,fig:challenges:red_filament,fig:challenges:red_stripe}).
Again, their presence hampers the detection task, making the recognition and the understanding of cells structure and size mandatory for the model.

A further source of complexity is represented by \textbf{overcrowding} (\Cref{fig:challenges:yellow,fig:challenges:green,fig:challenges:red_filament,fig:challenges:red_stripe}). When several cells are close-by, maybe partially overlapping, precisely localizing cell boundaries can be challenging, thus requiring adjustments to prevent the model from merging nearby cells into single agglomerations.

Last but not least, in some occasions the recognition of cells may be ambiguous even for human operators(cf. \textit{marked} and \textit{non-marked} instances in \Cref{fig:challenges:yellow,fig:challenges:green,fig:challenges:red_filament,fig:challenges:red_stripe}). Of course, this poses an issue of intrinsic \textbf{subjectivity} in the annotation process,
which in turn affects both the training and assessment phases.

By and large, all of these factors make the recognition and counting tasks harder and complicate the learning process.
Likewise, borderline annotations hinder model evaluation as their subjectivity deprives the model of a reliable and indisputable testbed.

\subsection*{Research lines}

As long as potential applications, the FNC dataset offers rich opportunities for diverse research directions, including:

\begin{itemize}
\item \textit{Object Segmentation, Detection, and Counting}: The dataset's comprehensive annotations and diverse neuronal structures support studies focusing on accurate segmentation, detection, and counting of cells. Particularly, FNC may be a challenging benchmark for class imbalance, object overlapping/overcrowding, and uncertainty estimation 

\item \textit{Transfer Learning}: With the availability of multiple image collections within FNC, researchers can explore transfer learning techniques, leveraging knowledge from one collection to improve performance on another.

\item \textit{Unsupervised or Self-/Weakly-Supervised Learning}: The presence of both labeled and unlabeled data within the FNC dataset provides an ideal testbed for evaluating unsupervised or self-/weakly-supervised learning approaches.

\item \textit{Evaluation of Annotation Types}: Researchers can investigate the effectiveness of different annotation types for specific tasks, allowing for a comparative analysis and selection of the most suitable annotations considering the cost/performance requirements of a given use-case.

\end{itemize}

\subsection*{Limitations}

Despite the Fluorescent Neuronal Cells collection presenting a variety of images in many aspects, it has limitations in terms of diversity across several parameters.

Firstly, all the images were collected by the same research laboratory in Bologna, utilizing fixed experimental conditions and acquisition settings. Furthermore, the images were captured using epifluorescence microscopy, which limits the range of techniques employed.
However, we believe that the adopted acquisition settings represent a more challenging scenario. Therefore, pre-training on FNC data should enable generalization to modern equipment such as confocal microscopy, which produces higher-quality images with sharper object boundaries and improved signal-to-noise ratio.

Another limitation lies in the lack of diversity in the cell types depicted and the animal species involved. Our dataset only focuses on subcellular components of rodent neurons. This might potentially impact the generalization of the models to different use cases and restrict their application to other cell types or animal species.

\section*{Code availability}

The code associated with this work is available on GitHub (\href{https://github.com/clissa/fluocells-scientific-data}{https://github.com/clissa/fluocells-scientific-data}).
The repository contains utils to:
\begin{itemize}
    \item perform \textbf{data operations} (\texttt{dataOps/}: i) converting raw \textit{TIFF} images into \textit{PNG} with metadata, ii) recreating expected data folders structure, iii) convert VIA annotation to binary masks, iv) encode binary masks into various annotation formats and types, v) preprocess yellow masks from previous FNC version\cite{clissa2021fluocells})
    \item implement deep learning \textbf{modelling} strategies (\texttt{fluocells/models/}: contains network blocks to implement c-ResUnet architecture\cite{morelli2021cresunet}; \texttt{compute\_metrics.py}, \texttt{evaluate.py} and \texttt{training.py}: contain utils to implement model training and evaluation)
    \item explore, analyze and evaluate models interactively (\texttt{notebooks/}: contains jupyter notebooks with examples of how to deal with standard stages of data analysis, namely i) exploratory data analysis, ii) implementation of model architecture and training pipeline, and iii) experiments 
\end{itemize}

\bibliography{sample}


\section*{Acknowledgements}

The study was supported with funding by the Italian National Institute for Nuclear Physics (INFN).
The collection of original images was supported by funding from the University of Bologna and the European Space Agency (Research agreement collaboration 4000123556).

\section*{Author contributions statement}

L.C. conceived the idea of curating and sharing the annotated data. 
M.L., R.A., M.C. and T.H. designed the biological study and M.L., A.O., E.P. and L.T. collected the original microscopic fluorescent images.
L.C. defined and orchestrated the annotation pipeline and instruments.
M.L., A.O., E.P. and L.T. performed data annotation and review.
L.C., A.M. and R.M implemented the code for data conversion, pre-processing, exploration, visualization and modelling.
L.C. and A.M. performed technical validation.
L.C. and A.M drafted the manuscript.
L.C., A.M., R.M., M.L., A.O., E.P. and L.T. proofread and revised the manuscript.
All authors read and approved the final manuscript.


\section*{Competing interests}

The authors declare no competing interests.

\end{document}